# Maintaining prediction quality under the condition of a growing knowledge space

Christoph Jahnz

ORCID: 0000-0003-3297-7564
Email: cjahnz@gmx.de

Intelligence can be understood as an agent's ability to predict its environment's dynamic by a level of precision which allows it to effectively foresee opportunities and threats. Under the assumption that such intelligence relies on a knowledge space any effective reasoning would benefit from a maximum portion of useful and a minimum portion of misleading knowledge fragments. It begs the question of how the quality of such knowledge space can be kept high as the amount of knowledge keeps growing. This article proposes a mathematical model to describe general principles of how quality of a growing knowledge space evolves depending on error rate, error propagation and countermeasures. There is also shown to which extend the quality of a knowledge space collapses as removal of low quality knowledge fragments occurs too slowly for a given knowledge space's growth rate.

## *A probabilistic dynamic model of knowledge spaces quality evolution*

In the age of modern epistemology *intelligence* is not an indisputable term as the concept of *reality* is neither. The term *intelligence* as used in following can accept a Glasersfeld's epistemological view of *constructivism* [1] as far as two conditions are accepted:

(I) An agent's intelligence enables it to derive regularities from experience in order to guide it to advantageous actions.

(II) There is an underlying reality which is consistent with itself, i.e. different perspectives on the same subject-matter do not result contradictorily [2].

This article further assumes an agent's *intelligence* is based on some *knowledge space*. A knowledge space is composed by knowledge fragments acquired during the agent's step by step learning. The knowledge space is a crucial part of the *world model* which the agent is establishing to gain orientation in its dynamic environment. Whether the knowledge space is understood, naively, as a representation of reality or whether it just composes the agent's subjective reality is not important in the context of this article as long as the basic assumptions (I) and (II) mentioned above are accepted.

In following, an atomic fragment of this knowledge space will be named *concept*. Inspired by constructivism, each of such concepts has been established by concepts already existing within an agent's knowledge space enriched by an agent's sensory information. New concepts are either assumed to be created by deductive [4][5][6] or by inductive [7][8] inference.

When an agent tries to predict its environment it usually selects a subset from its whole knowledge space due to some method. Using this knowledge subset, it deduces predictions about the environment.

However, the prediction quality depends on the quality of concepts used. The concepts can never be of perfect quality because mapping the world into a limited model means simplifying and losing information. A resulting prediction will probably be incorrect if relevant information has, for some reason, been lost, i.e. the prediction will conflict with the subsequently perceived result of the corresponding world behaviour.

Within our probabilistic model, the quality of each concept can be defined as an abstract *prediction success probability p ($p \in \mathbb{R}$, $0 \leq p \leq 1$)*, measuring how often a single concept contributes to a correct prediction result relative to the frequency of its invocation within any context. Within a concept space $\Omega$, the concepts with $p<P_{lim}$ ($P_{lim}\in\mathbb{R}$, $0\leq P_{lim}\leq 1$) will, by definition, be called *parasitic* and concepts with $p\geq P_{lim}$ will be called *accurate*, whilst $P_{lim}$ can be chosen arbitrarily.

Low quality concepts may be established for particular reasons such as overgeneralization [9] in inductive cognitive processes or processing failure or incomplete processing [10] during deductive cognitive processes. Such cognition failures can lead to the *miscreation* of a new concept. It is assumed that each new concept is formed as parasitic in consequence of erroneous logic or bad induction with a probability $P_{err}$ ($P_{err}\in\mathbb{R}$, $0\leq P_{err}\leq 1$).

Referring to the assumption, that new concepts usually are founded in existing ones, there can be identified still another possibility of a concept becoming parasitic: if at least one of its *base concepts* is parasitic (figure 1). A concept is defined as a *base concept* if it is necessary for the derivation of a child-concept by cognitive deduction [5].

As such, using the child-concept also means including the quality of all of its base concepts, giving a prediction success probability $p_{child} \leq min(p_{base(1)}, \ldots , p_{base(n)})$ where $p_{base}$ is the prediction success probability of the respective base concept and $n$ the number of all base concepts contributing to the child concept. In other words, the quality of a child-concept cannot be better than the worst quality of any of its base concepts. This way, quality problems are propagating throughout the concept space $\Omega$ by including base concepts of low quality.

Hierarchically structured pattern classifiers which help an intelligent agent to determine the semantic meaning of objects it is dealing with [3] are one typical example for deeply tree structured concept hierarchies including numerous base concepts for most of the concepts within those classifiers.

Assuming firstly a prediction success quality $p_n$ of any base concept for a specific child concept as statistically independent from $p_m$ of any other base concept and secondly that base concepts appear to be randomly chosen over $\Omega$, the resulting probability of forming a parasitic child-concept by including a parasitic base concept is

$$(1) \quad p_{ip} = 1 - (1 - \frac{c_p}{c})^B$$

where $c$ ($c\in\mathbb{R}_+$) is the total number of concepts in $\Omega$, $c_p$ ($c_p\in\mathbb{R}_+$) is the number of parasitic concepts in $\Omega$ and $B$ ($B\in\mathbb{N}$) is the number of base concepts for each child concept. Since further analysis considers a non-discrete growth process, the first two parameters are not defined as integers.

Assuming further that either the corresponding influence of $p_{ip}$ or $P_{err}$ is sufficient to cause a new concept to become parasitic, the total probability of a concept becoming parasitic results in

$$(2) \quad p_p = P_{err} + p_{ip} - P_{err} \cdot p_{ip} = 1 - (1 - P_{err})(1 - \frac{c_p}{c})^B \ .$$

Now the process of a growing concept space $\Omega$ can be described as

$$(3) \quad \Delta c_p = p_p \cdot L \cdot \Delta t - r_{cleanup} \cdot \Delta t$$

$$(4) \quad \Delta c = L \cdot \Delta t - r_{cleanup} \cdot \Delta t$$

where $\Delta c_p$ and $\Delta c$ ($\Delta c_p$, $\Delta c \in \mathbb{R}$, $0 < \Delta c_p$, $\Delta c \leq 1$) are the change rates of the respective concept counters for each time step $\Delta t$. $L$ describes the total number of new concepts generated each time step, and for simplicity, is assumed to be 1. This means, that each time step one new concept is generated.

The factor $r_{cleanup}$ ($r_{cleanup} \in \mathbb{R}_+$) describes a new aspect: Each time step there is not only created a potion of parasitic concepts determined by $p_p$, but it is also assumed that by some procedure each time step a portion of parasitic concepts is removed from $\Omega$ in order to cleanse the concept space of concepts with low prediction quality.

For now, with $L=1$ there can be derived the differential equation from (3), (4) and (2) as

$$(5) \quad \frac{\partial c_p}{\partial c} = p_p \cdot \frac{1}{1 - r_{cleanup}} - \frac{r_{cleanup}}{1 - r_{cleanup}}$$

which here will be named *equation of parasitic contamination*. It describes the evolution of the amount of parasitic concepts relative to the evolution of the total amount of concepts within an intelligent agent. Its solution provides us with an understanding of how contamination evolves when $c$ increases.

However, before solving the equation, $r_{cleanup}$ must be examined more closely. Two generic methods of identifying parasitic concepts will be introduced. Generally, it is assumed, that finding parasitic concepts is a non-trivial process within the intelligent agent. Otherwise the agent could have avoided establishing such concepts at creation time. Both clean-up approaches are based on statistical methods.

The first method proposes a clean-up by experience. This means that concepts are somehow sampled over time for their quality and if they are found to have insufficient quality, they will be removed from $\Omega$. One way would be to provide each concept with a cumulative quality rate, which is increased each time the concept participates in the satisfactory solution of some cognitive problem. Conversely, the cumulative quality rate is decreased if the solution is found to be unsatisfactory.

If the aggregate of quality rate stabilises below a certain limit, the concept will be removed. The general approach here is named *pragmatic reduction* and is modelled by

$$(6) \quad r_{prag} = \frac{c_p}{c} \cdot P_{prag\_ident} \cdot R_{prag\_freq} = \frac{c_p}{c} \cdot R_{prag}$$

where $\frac{c_p}{c} \cdot P_{prag\_ident}$ ($P_{prag\_ident} \in \mathbb{R}$, $0 \leq P_{prag\_ident} \leq 1$) represents the probability of identifying one parasitic concept by sampling during one time step and $R_{prag\_freq}$ ($R_{prag\_freq} \in \mathbb{R}_+$) represents the frequency by which this identification process occurs during one time step $\Delta t$. Both as constant assumed system parameters are combined to an abstract reduction parameter $R_{prag}$ which determines the performance of *pragmatic reduction*. Hereafter, $R_{prag}$ will be considered as independent of $c$.

The second method proposes a clean-up by competition. If there are present parasitic concepts, it is likely that $\Omega$ contains logical inconsistencies, since it is costly to check each concept at creation time for logical consistency against all concepts established in $\Omega$ [11]. Such inconsistencies will result in contradictory inferences about one and the same cognitive subject-matter [12]. Concepts which model reality well cause different inferences resulting consistently because – as a basic assumption in scientific theory – reality is assumed to be consistent with itself [2]. Thus, only any remaining concepts of low quality can be responsible for contradictions.

The clean-up method is supposed to locate contradictive concepts somehow and to identify the lower quality concepts by a success rate of statistic significance. This is similar to Darwinian selection as some *survival-of-the-fittest* method is used to decide about which of the contradictive concepts has to be removed from $\Omega$. The general approach here is named *competing reduction* and is defined by

$$(7) \quad r_{comp} = \frac{c_p}{c} \cdot \frac{c - c_p}{c} \cdot P_{comp\_match} \cdot P_{comp\_ident} \cdot R_{comp\_freq} = (\frac{c_p}{c} - (\frac{c_p}{c})^2) \cdot R_{comp}$$

where $\frac{c_p}{c} \cdot \frac{c - c_p}{c} \cdot P_{comp\_match}$ ($P_{comp\_match} \in \mathbb{R}$, $0 \leq P_{comp\_match} \leq 1$) represents the probability of one parasitic concept matching one contradictive accurate concept per time step $\Delta t$. $P_{comp\_ident}$ ($P_{comp\_ident} \in \mathbb{R}$, $0 \leq P_{comp\_ident} \leq 1$) represents the probability of the successful identification of such a parasitic concept during $\Delta t$ and $R_{comp\_freq}$ ($R_{comp\_freq} \in \mathbb{R}_+$) represents the frequency by which such identification process occurs during $\Delta t$. All different probabilities are assumed to be independent from each other. The three as constant assumed system parameters are combined to one abstract reduction parameter $R_{comp}$ which determines the performance of *competing reduction*. Hereafter, $R_{comp}$ will be considered as independent of $c$.

Finally, the independent contributions to parasitic reduction for each time step by (6) and (7) lead to

$$(8) \quad r_{cleanup} = r_{prag} + r_{comp} = R_{prag} \cdot \frac{c_p}{c} + R_{comp} \cdot (\frac{c_p}{c} - (\frac{c_p}{c})^2)$$

accomplishing the definition of all parameters contained within equation (5).

Different solutions for the evolution of parasitic contamination resulting from (5) for different system parameters are presented in figure 2 together with the results of a Monte Carlo simulation modelling the same scenario [13], but without using equation (5). The simulation was contributed in order to check if the probabilistic theory is correct. The results proved to be consistent as shown in figure 2.

The Monte Carlo simulation uses a *B* which is not constant for each concept but is, more realistically, chosen randomly with binominal distribution. As shown in figure 2, assuming the mean value of the binominal distribution $\overline{B}$ for solving (5) approximates fairly well the results of the Monte Carlo simulation within a certain range of *c*.

Now, we are able to answer the question, whether and if yes at what contamination $\frac{c_p}{c}$ the intelligent agent will stabilise depending on different sets of the constant parameters {$P_{err}$, $B$, $R_{prag}$, $R_{comp}$} for an ever-increasing *c*. Or, in other words, now we are able to calculate, what efforts of parasitic reduction are expected by our probabilistic model to preserve the concept space from becoming entirely contaminated by parasitic concepts.

Defining the contamination parameter $k \equiv \frac{c_p}{c}$ and $c > |\partial c|$ it can be proven [14] that

$$(9) \quad (1+R_{prag}+R_{comp})k - (R_{prag}+2R_{comp})k^2 + R_{comp}k^3 - 1 + (1-P_{err})(1-k)^B < 0$$

exists as condition for a growing contamination and

$$(10) \quad (1+R_{prag}+R_{comp})k - (R_{prag}+2R_{comp})k^2 + R_{comp}k^3 - 1 + (1-P_{err})(1-k)^B \geq 0$$

for a decreasing or constant contamination assuming an ever-increasing *c*.

It can be easily verified that (9) is given for *k=0* and $P_{err}>0$. This means that even contamination free concept spaces will start to become contaminated with a growing *Ω*.

As soon as the growing contamination *k* causes (10) to be fulfilled, the contamination stabilises at this *k*. Thus, the boundary contamination which is approached by the intelligent agent for an ever-growing *Ω* is determined by searching for the zero in (10) by starting at 0-contamination.

Conversely, setting a maximum contamination *k* at slightly below 1 and tracing down (10) until the first zero is encountered, will identify the *k* at which the intelligent agent will stabilise for an ever growing *Ω* after descending from a maximum contamination. If the initial contamination is total, i.e. equal to 1, the intelligent agent will remain stable at this contamination for an ever-growing *Ω*.

Figure 3 shows the final parasitic contamination for an infinitely growing *Ω* as function of *pragmatic* and *competing reduction* for different sets of {$P_{err}$, *B*}. For each configuration one scenario starts at an initial contamination of 0 and the other at nearly 1.

## *Interpretation and Conclusion*

Figure 3 shows a function type which for its characteristic shape here is named *plateau function*. Reducing the efforts for *pragmatic* and *competing reduction* sufficiently causes final *parasitic contamination* to approach towards a maximum of 1 which is on top of the plateau. Intelligent agents residing in such state will suffer from an ever decreasing quality within their knowledge space and therefore will be increasingly less able to make satisfying predictions about their environment.

The effect of pragmatic and competing reduction effort is asymmetric: If the effort for pragmatic reduction is critically low, not any level of competing reduction is able to improve knowledge space's quality evolution. In contrary, a sufficiently high pragmatic reduction effort is able to improve knowledge space's quality evolution even without any competing reduction effort.

As the threshold-like cliff edge of the plateau shows, there exists an area where increasing pragmatic or competing reduction effort only slightly will lead to a tremendous improvement of the knowledge space's quality evolution. Contrariwise, a slight decrease of either reduction will cause knowledge quality collapsing.

Interestingly, neither the amount of base concepts *B* nor the error probability $P_{err}$ significantly determine the quality evolution of the knowledge space. Instead, the quality evolution is mainly determined by the pragmatic and competing reduction efforts.

## *Outlook*

The mechanisms of concept space contamination mitigated by *pragmatic* and *competing reduction* seem to not be limited to the discipline of machine learning. Instead, they might be suspected to also play a role for learning in biological organisms and in human organisations. Further research might show whether there is any universality about this theory.

# **Figures**

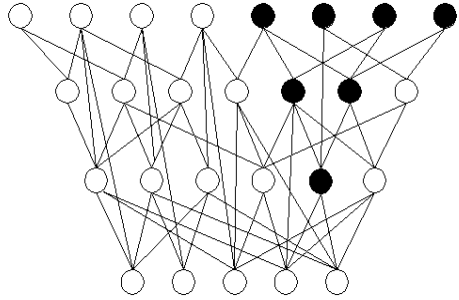

*Fig. 1:* parasitic propagation within a concept tree.

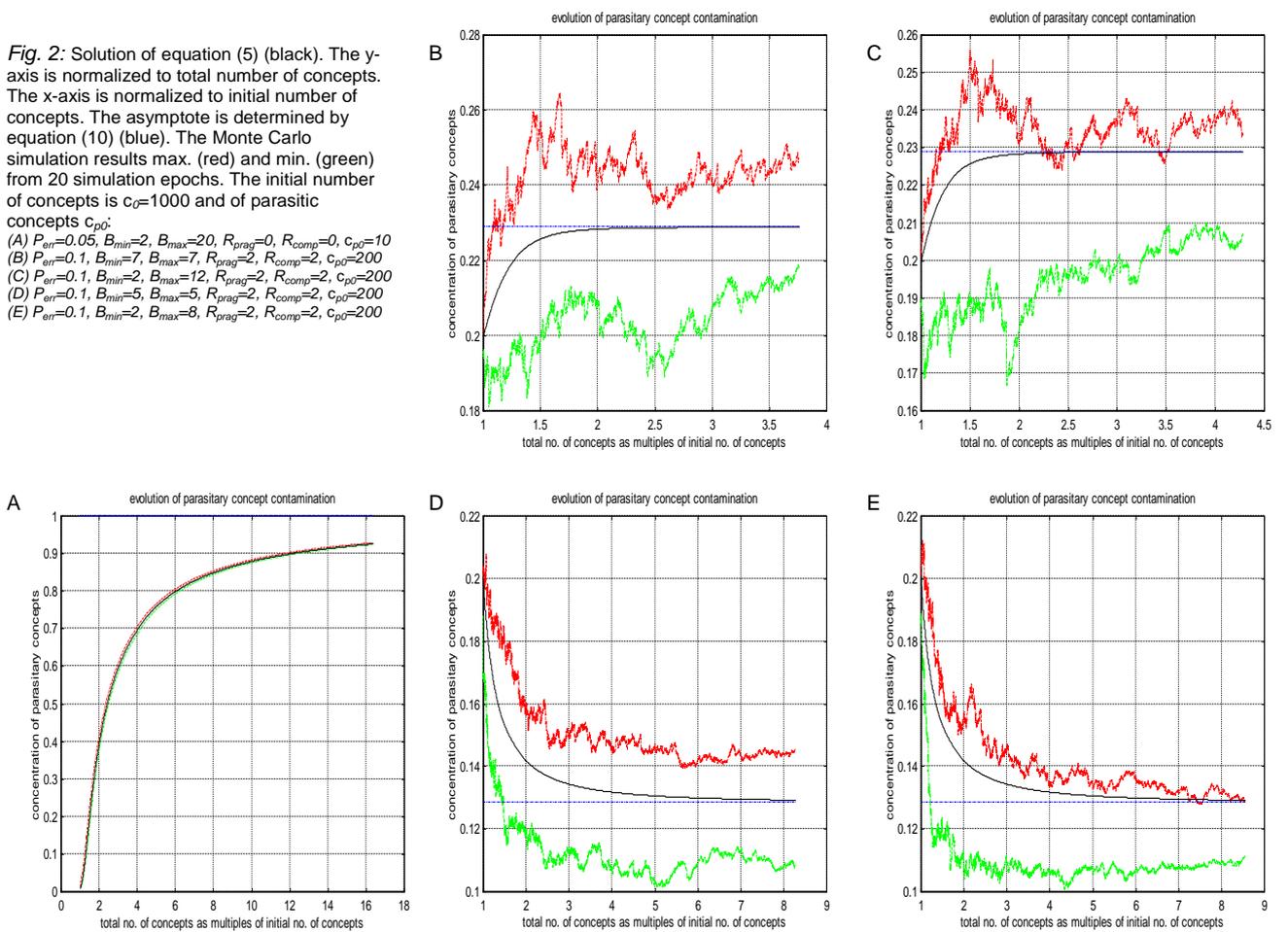

*Fig. 2:* Solution of equation (5) (black). The y-axis is normalized to total number of concepts. The x-axis is normalized to initial number of concepts. The asymptote is determined by equation (10) (blue). The Monte Carlo simulation results max. (red) and min. (green) from 20 simulation epochs. The initial number of concepts is $c_0$=1000 and of parasitic concepts $c_{p0}$:
(A) $P_{err}$=0.05, $B_{min}$=2, $B_{max}$=20, $R_{prag}$=0, $R_{comp}$=0, $c_{p0}$=10
(B) $P_{err}$=0.1, $B_{min}$=7, $B_{max}$=7, $R_{prag}$=2, $R_{comp}$=2, $c_{p0}$=200
(C) $P_{err}$=0.1, $B_{min}$=2, $B_{max}$=12, $R_{prag}$=2, $R_{comp}$=2, $c_{p0}$=200
(D) $P_{err}$=0.1, $B_{min}$=5, $B_{max}$=5, $R_{prag}$=2, $R_{comp}$=2, $c_{p0}$=200
(E) $P_{err}$=0.1, $B_{min}$=2, $B_{max}$=8, $R_{prag}$=2, $R_{comp}$=2, $c_{p0}$=200

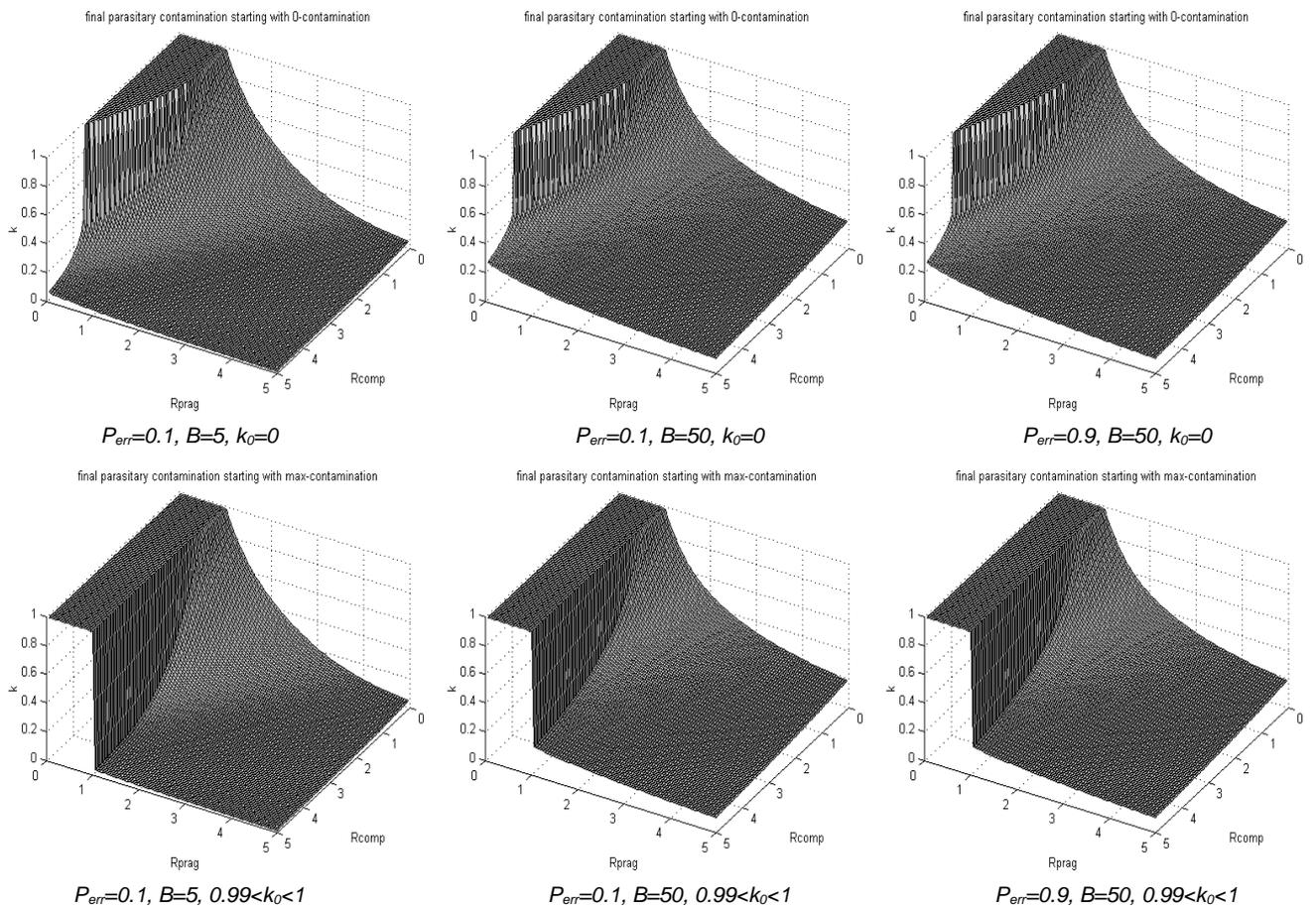

*Fig. 3:* Final parasitic contamination resulting from equation (10) with $k_0$ as initial parasitic contamination.